\documentclass[letterpaper]{article} 
\usepackage[preprint]{aaai2027}  
\usepackage[hyphens]{url}  
\usepackage{graphicx} 
\usepackage{natbib}  
\usepackage{caption} 
\usepackage{algorithm}
\usepackage{algorithmic}
\usepackage{amsmath}
\usepackage{amssymb}
\usepackage{svg}
\usepackage{subcaption}
\usepackage{newfloat}
\usepackage{listings}
\DeclareCaptionStyle{ruled}{labelfont=normalfont,labelsep=colon,strut=off} 
\floatstyle{ruled}
\newfloat{listing}{tb}{lst}{}
\floatname{listing}{Listing}

\usepackage{booktabs}

\title{AAAI Press Formatting Instructions \\for Authors Using \LaTeX{} --- A Guide}
\author{
    Written by AAAI Press Staff\textsuperscript{\rm 1}\thanks{With help from the AAAI Publications Committee.}\\
    AAAI Style Contributions by Peter Patel Schneider,
    Sunil Issar,\\
    J. Scott Penberthy,
    George Ferguson,
    Hans Guesgen,
    Francisco Cruz\equalcontrib\corresponding,
    Marc Pujol-Gonzalez\equalcontrib\corresponding
}
\affiliations{
    \textsuperscript{\rm 1}Association for the Advancement of Artificial Intelligence\\

    1101 Pennsylvania Ave, NW Suite 300\\
    Washington, DC 20004 USA\\
    proceedings-questions@aaai.org
}

\title{Learning Disease-Sensitive Latent Interaction Graphs From Noisy Cardiac Flow Measurements}
\author {
    Viraj Patel\textsuperscript{\rm 1}\corresponding,
    Marko Grujic\textsuperscript{\rm 2},
    Philipp Aigner\textsuperscript{\rm 2},
    Theodor Abart\textsuperscript{\rm 2},
    Marcus Granegger\textsuperscript{\rm 2},
    Deblina Bhattacharjee\textsuperscript{\rm 1},
    Katharine Fraser\textsuperscript{\rm 3}
}
\affiliations {
    \textsuperscript{\rm 1}Department of Computer Science, University of Bath, Bath, UK\\
    \textsuperscript{\rm 2}Christian Doppler Laboratory for Mechanical Circulatory Support, Department of Cardiac and Thoracic Aortic Surgery, Medical University of Vienna, Vienna, Austria\\
    \textsuperscript{\rm 3}Department of Mechanical Engineering, University of Bath, Bath, UK \\
    \{vbp24, db2466\}@bath.ac.uk, \{marko.grujic, philipp.aigner, theodor.abart, marcus.granegger\}@meduniwien.ac.at, khf27@bath.ac.uk
}

\begin{document}

\maketitle

\begin{abstract}
Cardiac blood flow patterns contain rich information about disease severity and clinical interventions, yet current imaging and computational methods fail to capture underlying relational structures of coherent flow features. We propose a physics-informed, latent relational framework to model cardiac vortices as interacting nodes in a graph. Our model combines a neural relational inference architecture with physics-inspired interaction energy and birth-death dynamics, yielding a latent graph sensitive to disease severity and intervention level. We first develop the method using computational fluid dynamics simulations of aortic coarctation, where learned interaction graphs reveal increasingly structured vortex interactions as vessel narrowing progresses. The resulting graph entropy exhibits a strong monotonic relationship with coarctation severity ($R^2=0.78$, Spearman $|\rho|=0.96$). We then evaluate the framework on fluid-structure interaction simulations of intracranial aneurysms using multiple geometric severity descriptors. Among these, aneurysm volume produces the most informative latent representation, with non-interaction graph entropy demonstrating a strong monotonic relationship with severity (Spearman $|\rho| = 0.91$) and generalising to several alternative morphological measures. Finally, we apply the approach to ultrasound-derived flow fields of a left ventricle under varying levels of left ventricular assist device support, where the latent graph captures the progressive loss of coherent vortex interactions under mechanical assistance, demonstrating cross-modal generalisation to imaging data. Across all datasets, latent interaction graphs and graph entropy provide interpretable markers of disease severity and intervention, linking haemodynamic organisation to clinically relevant physiological changes.
\end{abstract}


\section{Introduction}
Cardiovascular disease is the leading cause of death globally, and encompasses a diverse range of pathologies that alter blood flow, including congenital abnormalities such as coarctation of the aorta (CoA), cerebrovascular diseases such as intracranial aneurysms (IAs), and heart failure requiring left ventricular assist devices (LVADs) \cite{chongGlobalBurdenCardiovascular2025}. CoA, a narrowing of the aorta, is the most commonly misdiagnosed congenital heart disease in prenatal screening, though echocardiographic markers may improve detection rates \cite{dijkemaDiagnosisImagingClinical2017}. Issues inherent to echocardiography, like speckle noise and suboptimal Doppler angles, may cause underestimation of the severity of CoA \cite{bhattIsolatedCoarctationAorta2022}. Additionally, anatomical defects can be subtle in prenatal screening and can often mask pressure gradients and flow rate abnormalities \cite{villalainDiagnosticAccuracyPrenatal2024}. Patients with repaired CoA may develop left ventricular (LV) dysfunction with age, thus regular monitoring of the LV function is recommended from infancy through adulthood \cite{bhattIsolatedCoarctationAorta2022}. When intervention is required, an LVAD may be implanted \cite{hamadLVADTherapyCatalyst2023}. IAs, usually identified through phase-contract magentic resonance imaging (PC-MRI), arise from localised weakening of the cerebral arterial wall, resulting in vessel dilation and complex intra-aneurysmal flow disturbances associated with rupture risk \cite{sunderlandVortexAnalysisIntraAneurismal2016}. From prenatal screenings, to the implant of LVADs into patients, cardiovascular imaging plays a crucial role in diagnosis and monitoring.

Although denoising and despeckling of ultrasound images have been widely studied using both classical and deep learning approaches \cite{palhanoxavierdefontesRealTimeUltrasound2011, kaurCompleteReviewImage2023}, suboptimal scanning angles and occlusions remain difficult to recover \cite{osmanskiUltrafastDopplerImaging2012a}. Flow fields provide more sensitive functional information than raw images and are increasingly used in echocardiography, yet remain susceptible to noise, out-of-plane motion, and complex dynamics \cite{cantinottiCharacterizationAorticFlow2023, aignerVentricularFlowField2020, vignon-clementelEcoulementIntraventriculaireEchocardiographie2024}. While physics-informed reconstruction methods partially mitigate these issues, many deep learning models are still sensitive to ultrasound artefacts \cite{patelDynamicReconstructionUltrasoundDerived2025}. PC-MRI can suffer from incoherent velocities under complex and disturbed flow, and common geometrical parameters of IAs have led to inaccurate predictions of aneurysm rupture \cite{sunderlandVortexAnalysisIntraAneurismal2016}, suggesting that direct flow and shape measurements alone are insufficient for robust rupture prediction. An alternative is to represent flow structurally rather than pixel-wise. In particular, Gabriel graphs have been used to construct scale-invariant representations in which nodes correspond to critical points, like vortices, encoding the geometric organisation of the flow without relying on dense image information \cite{reza-soltaniRoleArtificialIntelligence2024, tobakTopologyThreedimensionalSeparated1981, kruegerQuantitativeClassificationVortical2019a}. Clinically, subtle and evolving haemodynamic defects motivate graph-based, scale-invariant representations that model the evolution and interaction of vortices.

We present a model that learns a scale invariant, noise-robust, physics-informed graph-based representation of various flow conditions. The main contributions are: (1) A neural relational inference model with a physics-head that learns disease-sensitive graph representations of cardiovascular flows, (2) The entropy of this learned representation as a quantitative measure of changes in vortex organisation associated with disease severity, (3) Validation of this model and marker across computational fluid dynamics (CFD) datasets of CoA, fluid-structure interaction (FSI) datasets of IAs and ultrasound scans of a heart under LVAD support.

\section{Background}
\subsection{Vortex Dynamics in Cardiovascular Disease}\label{sec:vortices}

Aortic flow in healthy individuals is predominantly laminar, exhibiting only a mild helical motion induced by curvature of the aortic arch \cite{gulanExperimentalStudyAortic2012}. In contrast, aortic coarctation disrupts this organised flow, producing high-velocity jets and complex vortical structures that can persist even after surgical repair due to long-term vascular remodelling \cite{hopeClinicalEvaluationAortic2010}. The narrowing generates a vortex ring downstream, which interacts with the aortic wall to produce secondary flow structures and increased flow complexity, altering the transport and mixing of blood throughout the descending aorta \cite{keshavarz-motamedEffectCoarctationAorta2014}. 

The literature demonstrates a strong link between aneurysm haemodynamics, morphology, and rupture risk. Complex vortex structures, particularly those interacting with the aneurysm wall, have been associated with aneurysm rupture, whereas most unruptured aneurysms exhibit comparatively simple flow patterns characterised by a single dominant, persistent vortex structure, with additional secondary vortices being transient and short-lived in comparison \cite{varbleIdentificationVortexStructures2017, HemodynamicMorphologicDiscriminants}. Morphological characteristics further influence these flow patterns, with larger aneurysm sac volumes corresponding to an increased risk of rupture \cite{varbleIdentificationVortexStructures2017}. Moreover, aneurysm shape has been shown to be a stronger predictor of rupture than size alone, as geometric features determine the capacity of the sac to support multiple persistent vortex structures, although such descriptors are more challenging to measure reliably in clinical practice \cite{dharMorphologyParametersIntracranial2008b}. In particular, sac height, the height-to-neck-width ratio (aspect ratio), and the surface-area-to-volume ratio (lobulation index) have all been reported to correlate significantly with rupture risk \cite{dharMorphologyParametersIntracranial2008b}.

Similar to CoA, as the left ventricle fills, the difference between the mitral valve and ventricle diameters causes a vortex ring, called the left ventricular vortex (LVV) \cite{pedrizzettiNatureOptimizesSwirling2005}. The formation time and geometry of the LVV are connected to heart failure and diastolic function \cite{gharibOptimalVortexFormation2006}. During diastole, the vortex becomes asymmetric as it travels towards the apex of the left ventricle due to colliding with the inclined wall \cite{kheradvarAssessmentTransmitralVortex2012}. This oblique vortex collision leads to the formation of a thick vortex arch with secondary vortex tubes rolled up around the primary one \cite{couchExperimentalInvestigationVortex2011, leVortexFormationInstability2012a}. When supported by an LVAD, flow is redirected through the apex of the left ventricle where the device is usually implanted. As LVAD support increases, the LVV circulation strength weakens as more blood flow is diverted through the device, which may significantly impact interactions with secondary flow structures \cite{aignerVentricularFlowField2020}.

\subsection{Neural Relational Inference}\label{sec:nri}
Given an interacting system of $N$ entities, let $\mathbf{x}_i^{(t)} \in \mathbb{R}^d$ be the feature vector of entity $i \in \{1,\cdots,N\}$ at timestep $t$. Neural Relational Inference (NRI) is an encoder-decoder architecture that learns a latent graph representation of the interacting system from a set of trajectories $\mathbf{x}_i = (\mathbf{x}_i^1, \cdots, \mathbf{x}_i^T)$ \cite{kipfNeuralRelationalInference2018}. In particular, the encoder is formalised by
\begin{equation}
    q_\phi(\mathbf{z}|\mathbf{x}) = \prod_{i \neq j} q_\phi(\mathbf{z}_{ij}|\mathbf{x}),
\end{equation}
where $\mathbf{z}_{ij} \in \mathbb{R}^{n_e}$ is the latent interaction vector between entities $i$ and $j$. This is trained with the decoder, $p_\theta(\mathbf{x}|\mathbf{z})$, to jointly optimise the ELBO
\begin{equation}
    \mathcal{L}(\phi,\theta) = \mathbb{E}_{q_\phi}[\log p_\theta(\mathbf{x}|\mathbf{z})] - \text{KL}[q_\phi(\mathbf{z}|\mathbf{x})||p(\mathbf{z})],
\end{equation}
where the first term minimises trajectory reconstruction error, and the second term ensures $\mathbf{z}_{ij}$ are meaningful. NRI has been proven effective on a wide range of problems, from physics simulations to protein interactions \cite{zhuNeuralRelationalInference2022}. Many modifications have been made to the original NRI architecture; notably dynamic NRI proposed by Graber et al. \cite{graberDynamicNeuralRelational2020} learned time-varying interactions $\mathbf{z}_{ij}(t)$. Yang et al. \cite{yangBehaviorInspiredNeuralNetworks2025} applied NRI to reinforcement learning problems and multi-agent systems, but found that current NRI architectures rely on the assumption that all interacting entities in the system are present at every timestep throughout the recorded trajectories. They expressed the need for a model that could account for missing data and the birth and death of an entity. In addition to this limiting assumption, NRI has yet to be adapted for systems with variable cardinality. When trained on datasets with many instances of a system, it is assumed that each instance has the same number of interacting entities.

\section{Method}
\subsection{Data Preparation}

\subsubsection{Flow Field Acquisition}
Flow fields for CoA were obtained from \cite{khanAortaPINNsData2025}. They created synthetic coarctations by narrowing a fixed region of the descending aorta, with severity defined as the percentage of the original radius (e.g. 40\% indicates the narrowed radius is 40\% of the baseline). They then performed transient 3D CFD simulations of pulsatile blood flow. To emulate ultrasound Echo-PIV conditions, a 2D cross-section was extracted with in-plane velocity projection, and multiplicative Rayleigh noise ($\sigma = 5,10$) was added \cite{karaogluRemovalSpeckleNoises2022}. 3D fluid-structure interaction (FSI) simulations of intracranial aneurysms \cite{goetzAnXploreComprehensiveFluidstructure2024} were preprocessed in the same way as the CoA dataset, and various severity measures were tested. Finally, LVAD data was acquired from an ex vivo mock circulatory loop using the explanted heart of a 55kg female sheep. The pulmonary vein, aorta, and pulmonary artery were connected to controlled reservoirs to regulate preload, afterload (Windkessel analogue), and coronary outflow, enabling simulation of varying LVAD support modes. Following configuration of support level, imaging was performed using a GE 6Vc-D echocardiographic probe. Echo-PIV is an ultrasound-based velocimetry technique that evaluates blood flow fields by tracking contrast microbubbles between successive image frames. In this study, microbubbles ($\sim 10\mu$m diameter) were injected into the bloodstream and their displacement was calculated via cross-correlation to generate 2D velocity vector fields \cite{crapperFlowFieldVisualization2000}. 

\subsubsection{Vortex Tracking}
Once 2D flow fields were obtained, vortices were identified in each frame using the SWIRL algorithm \cite{canivetecuissaInnovativeAutomatedMethod2022}, which applies the Rortex criterion to isolate rigid-body rotation while excluding shear effects, making it more robust than conventional swirling strength or vorticity methods for 2D flows. For each frame, a feature vector $\mathbf{x}_i^{(t)}$ (including the coordinate of centre, radius, orientation, existence $\in \{0,1\}$, and vorticity) of each vortex were recorded and compared between consecutive frames. Vortices in neighbouring frames that overlapped spatially, and had the same orientation, were considered to be the same vortex recorded at different timesteps. Trajectory data for all detected vortices were assembled into a single tensor, with zero-filled features at timesteps where a vortex was absent.

\subsection{Latent Interaction Model}
\begin{figure*}[tb]
    \centering
    \includegraphics[width=\linewidth]{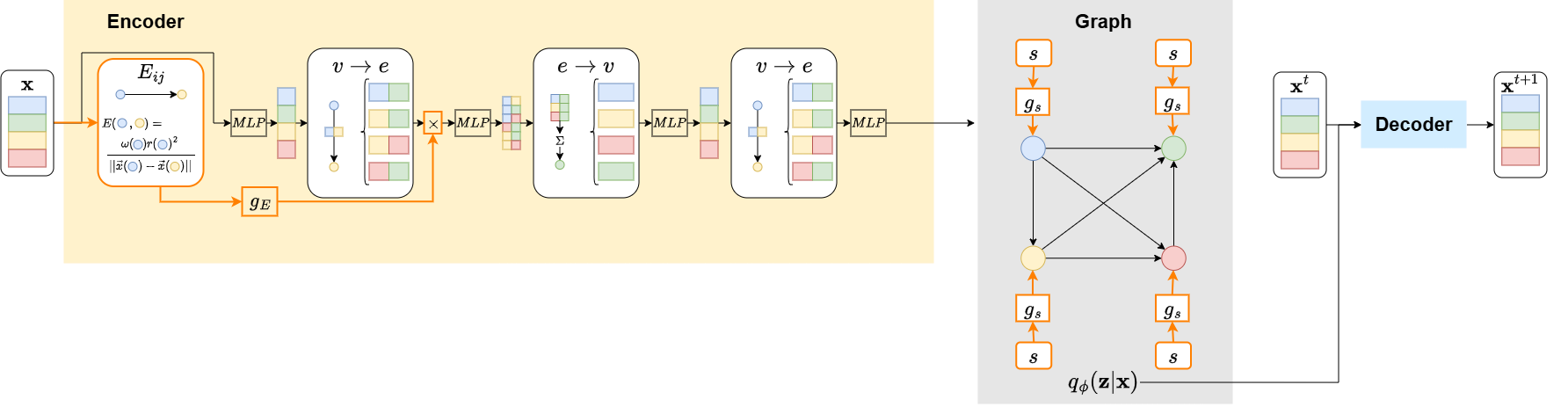}
    \caption{NRI architecture, with modifications compared with \cite{kipfNeuralRelationalInference2018} in orange. The encoder, which used a disease severity conditioned physics head, and decoder, are alternating MLPs and message passing layers. Solid blocks represent node-level embeddings and multi-coloured blocks represent edge-level embeddings.}
    \label{fig:nri}
\end{figure*}

Building on the original NRI architecture \cite{kipfNeuralRelationalInference2018}, several modifications were introduced to address the limitations outlined in the previous section (Figure \ref{fig:nri}, orange). A physics head $g_E$ was incorporated to encode circulation energy between vortices, and a conditioning network $g_s$ was added to account for coarctation severity, aneurysm measures, and LVAD support level. To accommodate variable vortex cardinality, a masked loss function was implemented during training. The latent relational structure retained two edge types (interaction/no interaction) with prior probabilities of 0.3 and 0.7, respectively.

\subsubsection{Energy Gating} 
For each pair of vortices, a proxy for the circulation energy inspired by the Biot-Savart Law \cite{perryTopologyFlowPatterns1994} was calculated
\begin{equation}
    E_{ij} = \frac{\omega_i r_i^2}{||\vec{x}_i - \vec{x}_j||_2},
\end{equation}
where $\omega_i$, $r_i$, and $\vec{x}_i$ are the vorticity, radius, and 2D coordinates of the centre of the $i$th vortex, respectively. This was chosen to be asymmetric to reflect the directed latent graph induced by temporal ordering. A single layer perceptron $g_E$ was used to map this interaction energy to the latent edge space. This allowed the encoder to learn how strongly the strength of the physical interaction should influence the inferred relational structure, rather than imposing a hard physical constraint.

\subsubsection{Conditioning on Severity}
The variable $s$ was used to encode severity. For the CoA dataset, this was the level of coarctation taking in a range of 30-100. For the IA dataset, the volume of the aneurysm sac was initially used as a measure of severity, with other measures also used for comparison. This introduced a different challenge of continuous severity values, as opposed to discrete levels in the other datasets. For the supported left ventricle, this was the level of support provided by the LVAD, with $0$ indicating no support, $0.5$ indicating partial support, and $1$ indicating full support. A linear layer $g_s$ conditioned the encoder on this severity and the result was multiplied by the logits output by the encoder, before being converted into the probability distribution $q_\phi$. This conditioning allowed for sensitivity of $s$ on the resulting latent interaction graph, without explicitly learning $s$.

\subsubsection{Temporal Consistency and Masked Loss}
The original NRI model used a placeholder for an adjacency matrix that prevents self-loops in the resulting latent graph \cite{kipfNeuralRelationalInference2018}. In our model, the entities were ordered by time of birth, and the graph placeholder ensured that later vortices could not have a causal effect on an earlier vortex, preserving temporal consistency. Due to a mixture of continuous and discrete features, different activation functions and loss functions were used to reconstruct each feature. We first define $x_o \in \{0,1\}^3$ as the one-hot encoded orientation and $x_e \in \{0,1\}$ as the binary value to determine if the vortex exists at a particular time. Given a feature vector $\mathbf{x} \in \mathbb{R}^8$, we define $\tilde{\mathbf{x}} := \mathbf{x} \setminus \{x_o, x_e\} \in \mathbb{R}^4$ as the vector of all continuous features. The reconstruction loss was masked by existence and defined as
\begin{equation}
    \mathcal{L}_\text{rec} = (\mathbb{E}_{q_\phi}[\log p_\theta(\tilde{\mathbf{x}}|\mathbf{z})] + CE(\hat{x}_o, x_o))x_e,
\end{equation}
where $CE(\cdot,\cdot)$ is the cross-entropy and $\hat{x}_o$ is the predicted orientation from the decoder. An existence loss $\mathcal{L}_\text{exist} = BCE(\hat{x}_e, x_e)$, where $BCE(\cdot,\cdot)$ is the binary cross entropy and $\hat{x}_e$ is the predicted value of existence from the decoder, was added to ensure that the model accurately predicted when a vortex exists. The total loss
\begin{equation}
    \mathcal{L} = \mathcal{L}_\text{rec} - \lambda_{KL}KL[q_\phi(\mathbf{z}|\mathbf{x})||p(\mathbf{z})] + \mathcal{L}_\text{exist}
\end{equation}
was jointly optimised by annealing $\lambda_{KL}$ (from 0 to 1 for CoA and LVAD datasets, 1 to 2 for aneurysm dataset) during the course of training.

\subsection{Implementation}
For fair comparison, the same training setup, including the train-test split ratio, proposed by Kipf et al. \cite{kipfNeuralRelationalInference2018} was used for this work. The CoA dataset contained 48 simulations, with 40 simulations used for training and 8 simulations used for testing and validation. The dataset was balanced such that each coarctation level and noise level was represented by the same number of simulations. The intracranial aneurysm (IA) dataset contained 105 simulations, with 84 used for training and 21 used for evaluation. The LVAD dataset contained 26 scans, with 18 scans used for training and 8 used for evaluating the performance of the model. This dataset contained 10 samples of no support, 8 samples of partial support, and 8 samples of full support. All instances of the model were trained on a NVIDIA GeForce RTX 3090 GPU with 24GB memory. The original NRI model was used as a baseline, with the performance of the model determined by the quality of reconstruction and the clinical significance of the resultant representation. The quality of reconstruction was measured by the mean squared error (MSE), as in \cite{kipfNeuralRelationalInference2018}, and mean absolute error (MAE) for all features except existence, which was measured by prediction accuracy. To measure the significance of the latent interaction graphs, with each edge-type treated as separate graphs, the entropy $H$ of each was computed and correlated with severity variable $s$. The $p$-value from Spearman correlation was recorded, in addition to a monotonicity (the fraction of consecutive severity-level pairs where entropy changes in the expected direction) score and the $R^2$ value of a least squares regressor between $H$ and $s$. Ethical approval from the relevant animal welfare committee was obtained prior to performing the experiments. The dataset and code will be released upon acceptance.

\section{Results and Discussion}
\subsection{CFD: Aortic Coarctation}\label{sec:coa_res}
To ensure performance was independent of parameter initialisation, the model was trained 5 times and mean and variance of each metric recorded. The existence accuracy ($0.7882$) and trajectory MSE ($0.0166$) were masked to avoid considering padded vortices. This MSE is low considering the range of the trajectory data, with small variance ($7.06\times10^{-6})$. The model predicted when vortices exist with high accuracy, and reconstructed trajectories retained a shape similar to the ground truth, albeit with some scaling and sign issues. Then the latent representation captured the structural patterns of the underlying vortex dynamics, with less emphasis on the exact trajectory magnitudes.

The distribution of probabilities assigned to the interaction edge-type varies with coarctation level (Fig. \ref{fig:coa_dist}): as the severity of the disease increases, the probability mass shifts toward stronger interactions, yielding an increase in latent graph entropy (Fig. \ref{fig:coa_dist}). The Spearman correlation coefficient of $-0.9553$ was consistent with this trend, with a highly significant $p$-value ($9.92\times10^{-4}$). Graph entropy explains a substantial proportion of disease severity variance ($R^2 = 0.7774$), indicating increasing interaction complexity as a dominant and systematic marker of disease progression rather than a weak or incidental correlation. This trend is explained by the underlying flow physics: as coarctation severity increases, the flow transitions toward more complex secondary structures, leading to a greater number and strength of vortex-vortex interactions. The relationship between entropy and radius is strongest when the radius is 60\% or smaller (Fig. \ref{fig:coa_ent}), likely because stronger confinement produces persistent wall-induced vortex interactions that increase coherence in the learned edge probabilities \cite{danailaFormationNumberConfined2018}.

\subsection{FSI: Intracranial Aneurysm}

The NRI model conditioned on aneurysm sac volume was trained using five random seeds, with metrics averaged across runs, using the entropy of the non-interaction graph for evaluation. The reconstruction MSE ($1.5592$) was higher than for the previously evaluated datasets due to the greater variability of aneurysm vortex dynamics. The aneurysm dataset contained fewer vortices on average but a large relative variation, leading to more variable birth-death dynamics and making trajectory reconstruction more challenging. This difficulty is compounded by the substantially different aneurysm sac geometries across cases, unlike the fixed geometries used in the CoA and LVAD datasets, and is reflected in the lower but still reasonable existence accuracy of $0.6893$. Despite this, the learned graph representations remained highly informative with non-interaction graph entropy decreasing as aneurysm volume increased (Fig. \ref{fig:aneurysm_dist} and \ref{fig:aneurysm_ent}), achieving a Spearman correlation of $\rho=-0.9053$ with a corresponding $p$-value of $1.7275\times10^{-7}$, with all five runs achieving statistical significance, and a high monotonicity score of $0.8705$. Although the linear $R^2$ was relatively low ($0.0875$), this is consistent with a nonlinear relationship between entropy and aneurysm sac volume. Increasing aneurysm volume promotes the formation of larger, more spatially isolated recirculation regions, reducing the diversity of interaction patterns between vortices and consequently lowering the entropy of the non-interaction graph.

In addition to aneurysm volume, six clinically relevant geometric descriptors were investigated as alternative conditioning variables: maximum diameter, sac height, neck width (NW), aspect ratio (AR), lobulation index (LI), and non-sphericity index (NSI). These measures characterise complementary aspects of aneurysm morphology and are commonly associated with rupture risk \cite{dharMorphologyParametersIntracranial2008b}. Their suitability as conditioning variables was evaluated in the following experiments.
Table \ref{tab:fsi} demonstrates that the choice of conditioning variable is critical for learning meaningful latent graph representations. Conditioning on aneurysm volume consistently outperformed all other severity measures, achieving the strongest Spearman correlation, highest monotonicity, and significant correlations across all five training runs. In contrast, diameter and sac height produced only moderate monotonic relationships ($\rho \approx -0.47$, monotonicity $\approx 0.67$) with significance in only three of five runs, while the others and the unconditioned model failed to produce consistent or statistically significant correlations. Although sac height achieved the highest $R^2$ (0.3216), this reflects the application of a linear regressor to a predominantly nonlinear relationship and is therefore less representative than the rank-based metrics. Overall, these results indicate that the conditioning variable is not interchangeable; aneurysm volume provides the most physically informative conditioning signal for organising vortex interactions into latent representations that capture disease sensitivity.

\begin{table*}[tb]
\caption{Performance of models conditioned on different aneurysm severity measures, averaged over five random seeds.}\label{tab:fsi}
\centering
\begin{tabular}{lccccc}
\toprule
\multicolumn{1}{c}{\textbf{Severity Measure}} & 
\textbf{$\rho$} & 
\textbf{$p$-value} & 
\textbf{$R^2$} &
\textbf{Monotonicity} &
\textbf{No. Signficant Runs /5} \\ 
\cmidrule(lr){1-1}\cmidrule(lr){2-2}\cmidrule(lr){3-3}\cmidrule(lr){4-4}\cmidrule(lr){5-5}\cmidrule(lr){6-6}
Volume & \textbf{-0.9053} & $\mathbf{1.7275\times10^{-7}}$ & 0.0875 & \textbf{0.8705} & \textbf{5} \\
Diameter & -0.4758 & 0.0955 & 0.2531 & 0.6733 & 3 \\
Sac Height & -0.4696 & 0.0740 & \textbf{0.3216} & 0.6743 & 3 \\
Neck Width & 0.1229 & 0.6209 & 0.0711 & 0.4619 & 0 \\
Aspect Ratio & -0.1005 & 0.4092 & 0.1015 & 0.5362 & 1 \\
Lobulation Index & 0.0086 & 0.4725 & 0.0985 & 0.4886 & 1 \\
Non-sphericity Index & 0.0078 & 0.6452 & 0.0178 & 0.4952 & 0 \\
Unconditioned & 0.0184 & 0.6185 & 0.0179 & 0.4962 & 0 \\ \bottomrule
\end{tabular}
\end{table*}

\begin{figure}[h]
    \centering
    \includegraphics[width=\linewidth]{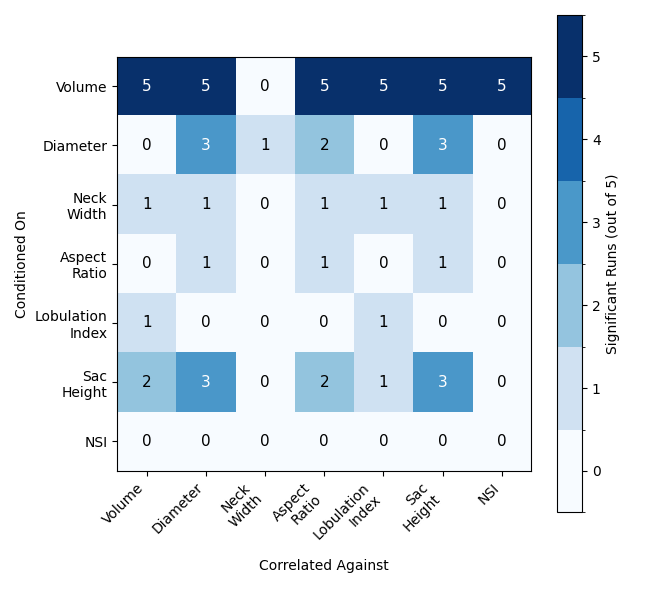}
    \caption{Cross-evaluation of conditioned models against alternative aneurysm severity measures. The entry in each cell represents the number of statistically significant Spearman correlations ($p<0.05$) between graph entropy and the corresponding severity measure across five random seeds.}
    \label{fig:fsi_conditioning}
\end{figure}

To investigate whether the learned representation generalises beyond the conditioning variable, each conditioned model was correlated against all alternative severity measures. Fig. \ref{fig:fsi_conditioning} shows that models conditioned on aneurysm volume consistently produced significant correlations with all other severity measures except neck width, whereas models conditioned on alternative descriptors exhibited substantially weaker and less reproducible relationships. Despite the strong correlations between several geometric measures, these findings demonstrate that the conditioning variable is not interchangeable. Instead, aneurysm volume uniquely organises vortex interactions into a latent representation that captures multiple aspects of aneurysm morphology. The ability of graph entropy to generalise beyond the conditioning variable suggests that the learned representation captures broader haemodynamic information associated with multiple aneurysm severity measures, potentially complementing geometric measurements that are more difficult to obtain reliably in clinical practice.

\subsection{Ultrasound: LVAD Supported Ventricle}

When extended to ultrasound scans of an LVAD supported left ventricle, the existence accuracy and MSE of the reconstructed trajectories were $0.7604$ and $0.0039$ respectively. In contrast with the CoA dataset, the reconstruction maintained the same sign as the ground truth data but suffered similar scaling issues. The model's prediction of vortex existence timing aligned almost perfectly with the ground truth.

\begin{figure*}[tb]
    \centering
    \begin{subfigure}[b]{0.45\linewidth}
        \centering
        \includegraphics[width=\linewidth]{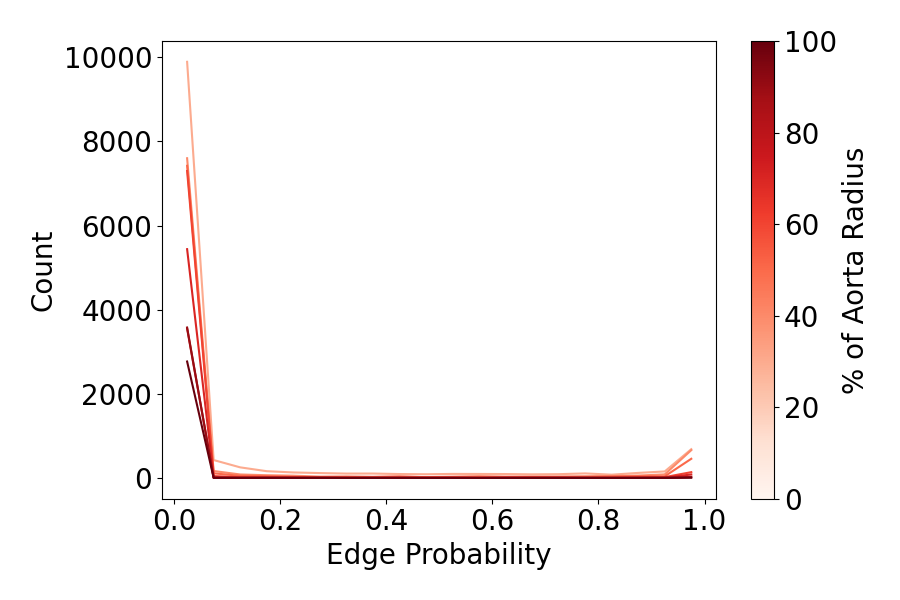}
        \caption{}
        \label{fig:coa_dist}
    \end{subfigure}
    \begin{subfigure}[b]{0.45\linewidth}
        \centering
        \includegraphics[width=\linewidth]{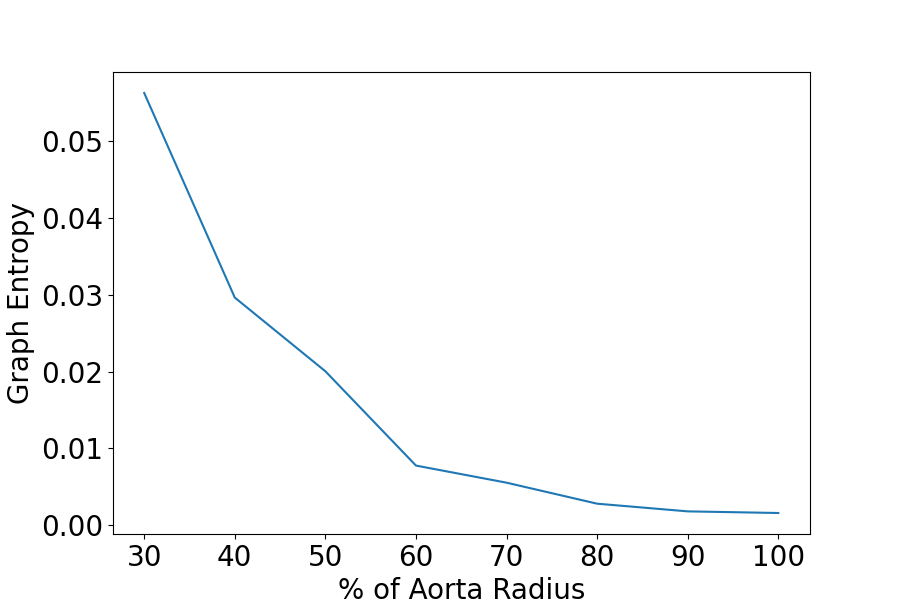}
        \caption{}
        \label{fig:coa_ent}
    \end{subfigure}
    \begin{subfigure}[b]{0.45\linewidth}
        \centering
        \includegraphics[width=\linewidth]{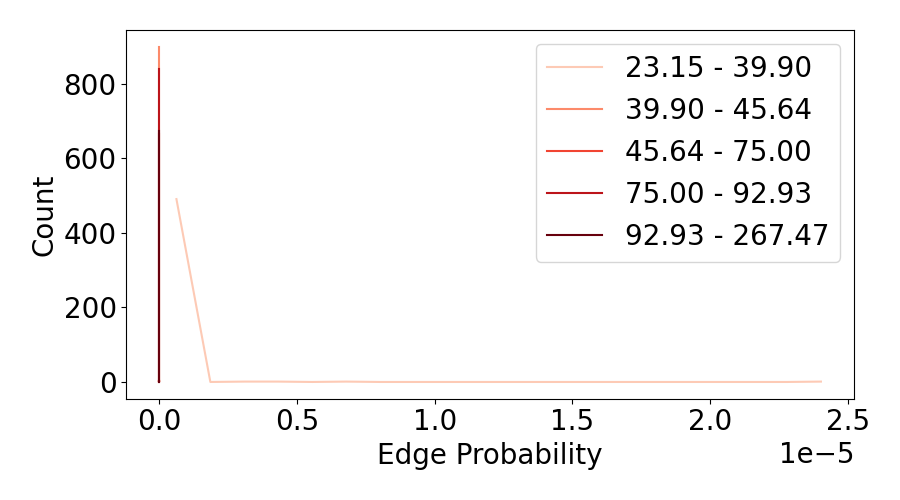}
        \caption{}
        \label{fig:aneurysm_dist}
    \end{subfigure}
    \begin{subfigure}[b]{0.45\linewidth}
        \centering
        \includegraphics[width=\linewidth]{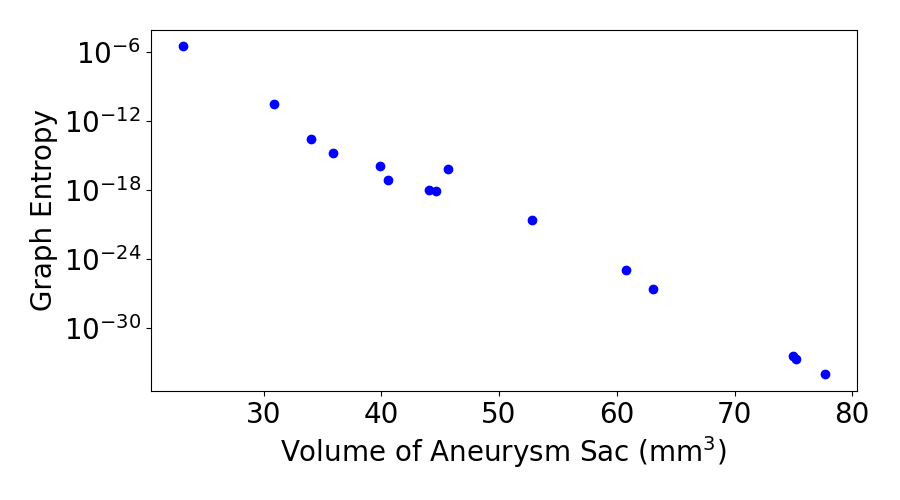}
        \caption{}
        \label{fig:aneurysm_ent}
    \end{subfigure}
    \begin{subfigure}[b]{0.45\linewidth}
        \centering
        \includegraphics[width=\linewidth]{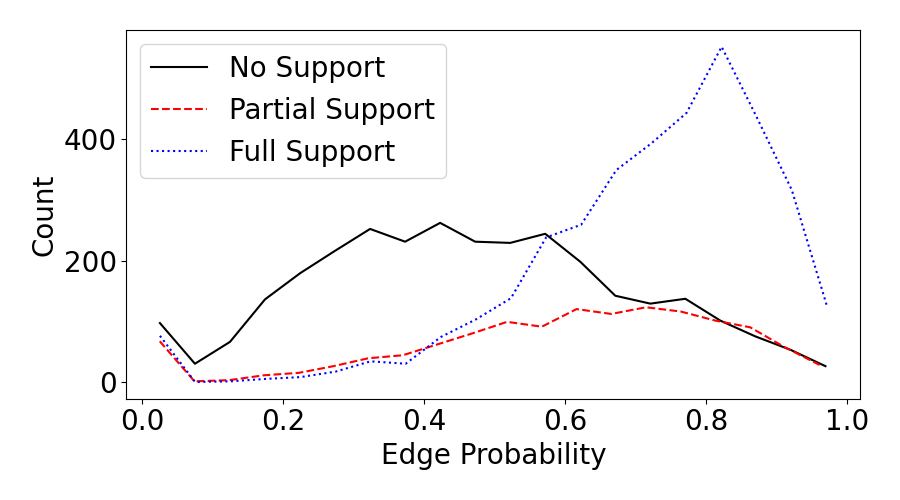}
        \caption{}
        \label{fig:lv_dist}
    \end{subfigure}
    \begin{subfigure}[b]{0.45\linewidth}
        \centering
        \includegraphics[width=\linewidth]{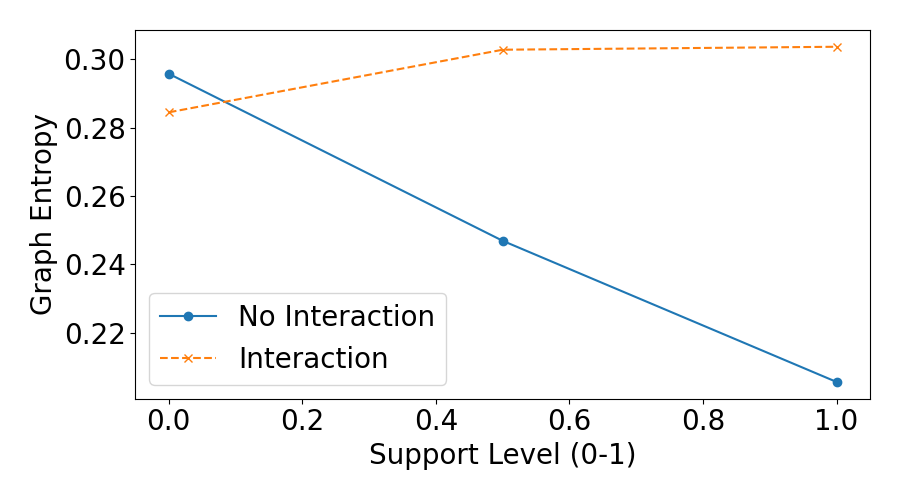}
        \caption{}
        \label{fig:lv_ent}
    \end{subfigure}
    \caption{Aortic coarctation: (a) the distribution of edge probability and (b) the entropy of the latent interaction graph changes with coarctation level. Intracranial aneurysm: (c) the distribution of edge probability for binned aneurysm volumes (no interaction) and (d) the entropy of the latent graph changes with aneurysm volume. LVAD support: (e) the distribution of edge probability (for no interaction) and (f) the entropy of the latent interaction graph changes with LVAD support level. Decreasing \% of aorta radius corresponds to increasing disease severity. LVAD support levels: 0.0 = no support, 0.5 = partial support, and 1.0 = full support.}
    \label{fig:lv_rep}
\end{figure*}

When investigating the latent representation, it was found that the first edge type, corresponding to no causal interaction, was more indicative of the change in flow conditions. In particular, as the support level increases, the distribution of edge weights skews to higher values with lower variance (Fig. \ref{fig:lv_dist}). This is in agreement with the decrease in entropy of the non-interaction latent graph with increasing support level (monotonicity score of $1.0$, Fig. \ref{fig:lv_ent}). Spearman correlation produced a coefficient of $-0.9449$ with a $p=0.0004$, and $R^2=0.8549$. These results imply that as the support level increased, many vortex pairs become consistently non-interacting and the model becomes increasingly confident about this across different scans. In contrast to the non-interaction graph, the entropy of the interaction graph increases between no support and partial support, but plateaus between partial and full support (Fig. \ref{fig:lv_ent}). 

This suggests a transitional regime in which residual vortex interactions weaken before stabilising or disappearing at higher support. This can be explained by competing inflow-outflow pathways: natural filling ejects flow through the aortic valve, whereas full support redirects it towards the apex \cite{wongIntraventricularFlowPatterns2014}, and partial support strips vorticity into diffuse, less coherent structures \cite{ghodratiValidationNumericallySimulated2021}.

\subsection{Ablations}

Our full model achieves the lowest MAE, reflecting improved average reconstruction of vortex trajectories, while the ablation with no severity conditioning has slightly higher MAE but lower MSE, indicating fewer extreme deviations (Table \ref{tab:abl} reports best of 5 training runs; mean results given in previous section). Ablations reveal that vortex ordering improved existence accuracy and reconstruction by providing temporal context for vortex births. Moreover, physics gating enhanced the latent representation by incorporating circulation energy constraints that capture physically meaningful interactions. Finally, severity conditioning allowed the model to specialise across disease levels, producing representations that more accurately reflect the flow alterations induced by coarctation.

\begin{table*}[tb]
\caption{Ablations on NRI model applied to CoA dataset.}\label{tab:abl}
\centering
\scalebox{1.0}{
\begin{tabular}{lccccccc}
\toprule
\multicolumn{1}{c}{\textbf{Ablation}} &
\textbf{\begin{tabular}[c]{@{}c@{}}Existence\\ Accuracy\end{tabular}} &
\textbf{MAE} &
\textbf{MSE} &
\textbf{$\rho$} &
\textbf{$p$-value} &
\textbf{$R^2$} &
\textbf{\begin{tabular}[c]{@{}c@{}}Monotonicity\end{tabular}} \\
\cmidrule(lr){1-1}\cmidrule(lr){2-2}\cmidrule(lr){3-3}\cmidrule(lr){4-4}%
\cmidrule(lr){5-5}\cmidrule(lr){6-6}\cmidrule(lr){7-7}\cmidrule(lr){8-8}
None
& $\mathbf{0.840}$
& $\mathbf{0.013}$
& $0.015$
& $\mathbf{-0.988}$
& $\mathbf{4.26\times10^{-6}}$
& $\mathbf{0.915}$
& $\mathbf{1.00}$ \\
No ordering
& $0.605$
& $0.014$
& $0.015$
& $-0.691$
& $0.058$
& $0.845$
& $0.800$ \\
No physics gating
& $0.778$
& $0.014$
& $0.016$
& $-0.167$
& $0.693$
& $0.000$
& $0.640$ \\
No severity conditioning
& $0.704$
& $0.013$
& $\mathbf{0.011}$
& $0.518$
& $0.188$
& $0.418$
& $0.308$ \\
Original NRI
& $0.617$
& $0.016$
& $0.019$
& $0.458$
& $0.254$
& $0.210$
& $0.308$ \\
\bottomrule
\end{tabular}
}
\end{table*}

\begin{figure}
    \centering
    \includegraphics[width=\linewidth]{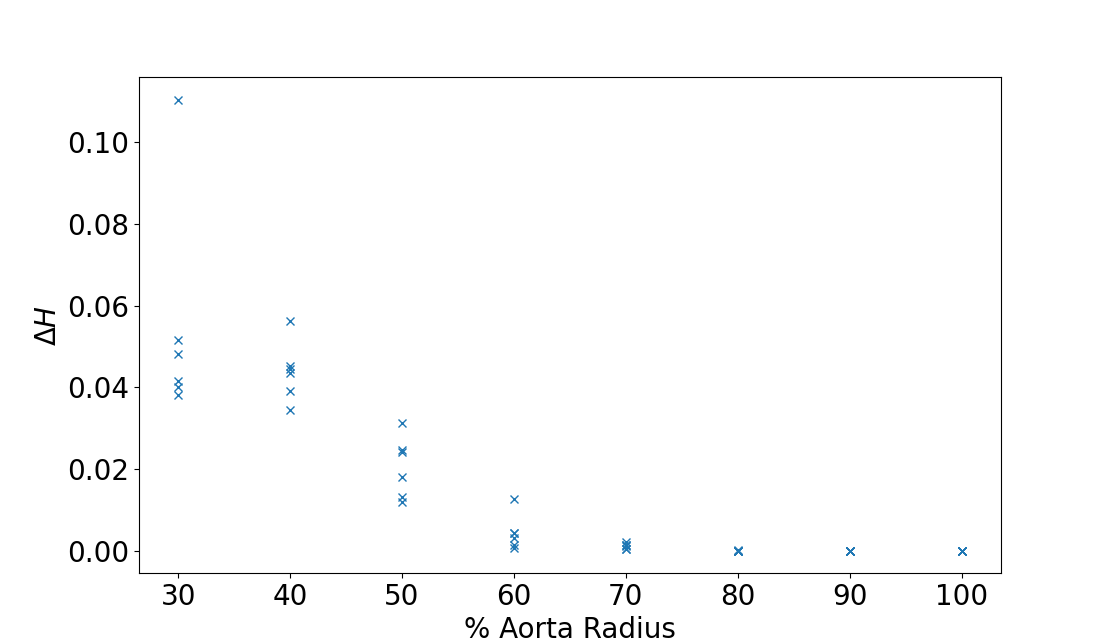}
    \caption{Change in graph entropy for perturbations of $\pm 10$ to severity for the CoA dataset. $\Delta H$ was largest at severe coarctation and lowest in healthier regimes.}
    \label{fig:abl_pert}
\end{figure}

Although the asymmetry in Fig. \ref{fig:fsi_conditioning} shows that severity conditioning does not introduce a trivial $s \to H$ mapping, to further verify it, we performed a perturbation analysis: for each CoA simulation, $s$ was varied by $\pm 10$  and the resulting change in entropy $\Delta H$ was recorded. Fig. \ref{fig:abl_pert} shows how the sensitivity is strongly state-dependent: $\Delta H$ is largest at severe coarctation (30-40\% radius) and diminishes to near zero at mild levels (80-100\%), confirming that the conditioning modulates the encoder's response to flow features rather than memorising a fixed mapping.

\subsection{Limitations and Future Work}
The datasets used in this study remain modest in size, and external validation on independent cohorts is required to establish generalisability. Increasing the diversity of aortic geometries, acquiring echocardiographic data from additional subjects, and evaluating a larger cohort of aneurysm geometries with clinical outcomes would further strengthen confidence in the learned representations. Furthermore, the current binary edge formulation may oversimplify subtle interactions, as vortex coupling strength is continuous in reality. Future work will explore continuous relational representations and additional physical constraints to more accurately capture the evolution of haemodynamic interactions.

\section{Conclusion}

We introduced a noise-robust, physics-informed neural relational inference framework to characterise vortex interaction structure in altered cardiovascular flow conditions, including coarctation of the aorta, intracranial aneurysms, and LVAD-supported left ventricular flow. By analysing the entropy of inferred edge types, we identified systematic changes in relational uncertainty associated with increasing vascular obstruction severity, aneurysm morphology, and mechanical support. In coarctation, relational entropy captured progressive disruption of flow structures as the aortic radius narrowed. In intracranial aneurysms, decreasing entropy of the non-interaction edge type with increasing aneurysm volume indicated that larger sacs were associated with more organised vortex interaction patterns, while correlations between graph entropy and alternative morpohological descriptors demonstrated that the learned representation generalised beyond the conditioning variable. In the LVAD setting, decreasing entropy of the non-interaction edge type alongside a transient increase in interaction entropy suggested a transitional regime at intermediate support levels, reflecting competition between native pulsatile dynamics and pump-driven flow. Together, these results demonstrate that relational modelling provides a compact and interpretable representation of complex haemodynamic flow structures, highlighting its potential as a general framework for studying pathological cardiovascular flow.

\bibliography{representations}


\end{document}